\documentclass[10pt,twocolumn,letterpaper]{article}
\pdfoutput=1
\usepackage{wacv}
\usepackage{times}
\usepackage{epsfig}
\usepackage{graphicx}
\usepackage{amsmath}
\usepackage{amssymb}

\def\wacvPaperID{414} %

\wacvfinalcopy %

\ifwacvfinal
\def\assignedStartPage{1} %
\fi

\ifwacvfinal
\usepackage[breaklinks=true,bookmarks=false]{hyperref}
\else
\usepackage[pagebackref=true,breaklinks=true,colorlinks,bookmarks=false]{hyperref}
\fi

\ifwacvfinal
\else
\fi

\newcommand{\R}{\mathbb{R}}
\newcommand{\beginsupplement}{%
        \setcounter{table}{0}
        \renewcommand{\thetable}{S\arabic{table}}%
        \setcounter{figure}{0}
        \renewcommand{\thefigure}{S\arabic{figure}}%
        \setcounter{section}{0}
        \renewcommand{\thesection}{S\arabic{section}}%
        \clearpage
        \pagebreak
     }

\begin{document}

\title{Modality-Agnostic Attention Fusion for visual search with text feedback}

\author{Eric Dodds, Jack Culpepper, Simao Herdade, Yang Zhang\\
Yahoo Research\\
{\tt\small \{edodds, jackcul, sherdade, yang.zhang\}@verizonmedia.com } \\
\and
Kofi Boakye\\
Pinterest\\
{\tt\small kofi@pinterest.com}
}

\maketitle
\begin{abstract}
Image retrieval with natural language feedback offers the promise of catalog search based on fine-grained visual features that go beyond objects and binary attributes, facilitating real-world applications such as e-commerce. Our Modality-Agnostic Attention Fusion (MAAF) model combines image and text features and outperforms existing approaches on two visual search with modifying phrase datasets, Fashion IQ and CSS, and performs competitively on a dataset with only single-word modifications, Fashion200k. We also introduce two new challenging benchmarks adapted from Birds-to-Words and Spot-the-Diff, which provide new settings with rich language inputs, and we show that our approach without modification outperforms strong baselines. To better understand our model, we conduct detailed ablations on Fashion IQ and provide visualizations of the surprising phenomenon of words avoiding ``attending'' to the image region they refer to.
\end{abstract}

\section{Introduction}
In recent years, the task of image retrieval has advanced to allow for more complex, multi-modal queries, including those involving both images and text \cite{guo2018dialog,guo2019fashion,vo2019composing, Hosseinzadeh_2020_CVPR, Chen_2020_CVPR}.
Such systems accommodate users by offering a more expressive framework for conveying image search targets.

Here we study the setting in which a query is specified by an image together with text that describes how to alter the image to arrive at the desired concept.
We are particularly interested in distinguishing fine-grained differences between images, which may not be characterized by the presence, absence, or location of discrete objects. This setting includes valuable applications such as searching a fashion catalog, and it is a good match for the flexibility of natural language compared to pre-specified lists of attributes or sets of objects.

\begin{figure}
\begin{center}
\includegraphics[width=0.48\textwidth]{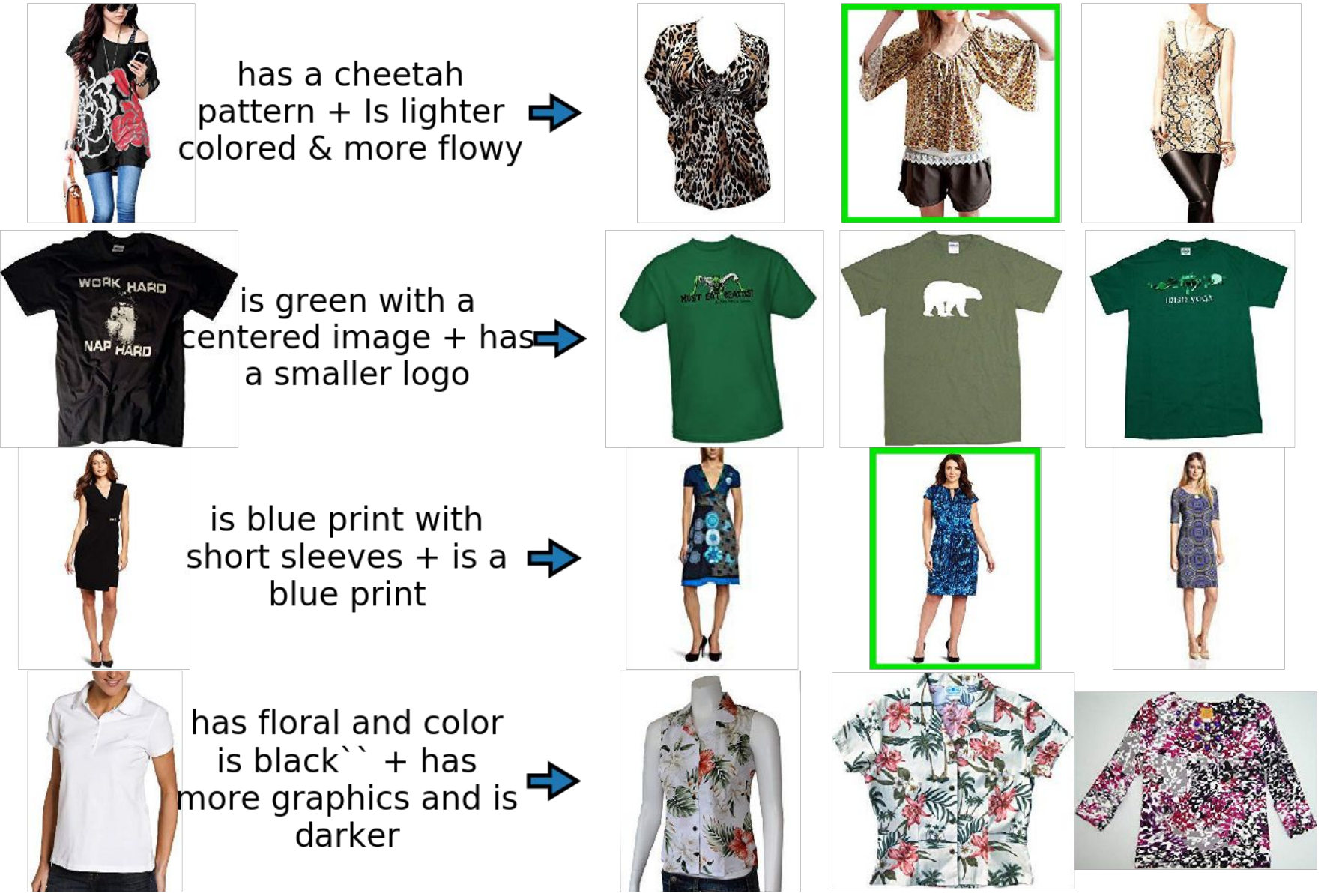}
\end{center}
   \caption{Retrieval examples on the Fashion IQ validation set. A query image and two modifying captions are provided, here joined by +. A green rectangle indicates the labeled target image. The dataset is not exhaustively labeled.}
\label{fig:fiq_examples}
\end{figure}

Vo et al. \cite{vo2019composing} introduced datasets to study this task and demonstrated that combining text and image features with a residual gating mechanism yielded the best results among adaptations of several existing methods. More recent advances have used attention mechanisms~\cite{Hosseinzadeh_2020_CVPR, Chen_2020_CVPR} and introduced the Fashion IQ dataset, providing a setting with richer text modifications~\cite{guo2019fashion}.

We propose a simple model we call Modality-Agnostic Attention Fusion (MAAF) to address the text-modified image retrieval task. The core fusion idea is to extract ``token" vectors representing parts of the input for each modality, then stack these tokens into a single sequence to process with an attention model.
Applying this broad idea to our specific task requires extracting useful tokens from text and from images that may not have multiple objects, and finally combining the attention model outputs into a single vector suitable for nearest-neighbor search. We compare approaches for each of these steps and show that a simple LSTM for the text input, convolutional feature maps at multiple scales for the image input, and average pooling within each class of token achieves the strongest results on Fashion IQ. 
We hypothesized that our flexible modality-agnostic approach would outperform two-streamed approaches with cross-modal attention, and we provide experiments to confirm this hypothesis on Fashion IQ.

Our best model outperforms the state of the art on Fashion IQ as well as CSS3D~\cite{vo2019composing}, and we show competitive results on Fashion-200k and MIT States. These latter datasets have only single-word text modifications, so we also introduce two new benchmarks with richer language components. These benchmarks are based on Birds-to-Words~\cite{forbes2019neural} and Spot-the-Diff~\cite{jhamtani2018learning}, and we show that our model outperforms strong baselines including TIRG~\cite{vo2019composing} on them without modification.

We also show through visualizations the counter-intuitive behavior of words \textit{avoiding} attending to the regions they refer to. To the best of our knowledge, this work is the first to identify such behaviour, which we attribute to the nature of the task.

In brief, we make the following contributions:
\begin{itemize}
    \item We present Modality-Agnostic Attention Fusion (MAAF), an approach to image retrieval using attention over undifferentiated text and image queries, which yields improvements over the state-of-the-art in several retrieval tasks.
    \item We quantitatively demonstrate the utility of various components of the retrieval system through baseline and ablation experiments.
    \item We introduce two new challenging benchmarks for evaluating models on the multimodal image retrieval task.
    \item We show through visualizations that our attention mechanism exhibits the counter-intuitive behavior of attending to complementary regions of those the modifying sentence word tokens refer to. 
\end{itemize}

\section{Model}
\begin{figure*}
\begin{center}
\includegraphics[width=\textwidth]{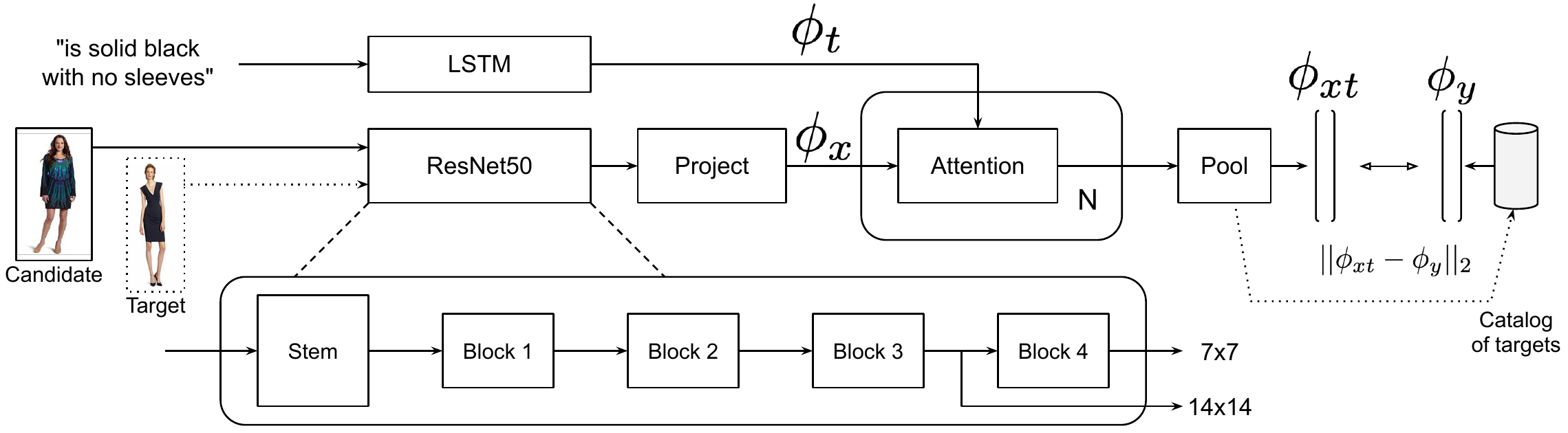}
\end{center}
   \caption{Illustration of MAAF. A candidate image is indicated by the solid outline, and a target image is indicated by the dotted outline. Image tokens are extracted from blocks 3 and 4 of ResNet, then projected to 512 dimensions and flattened into a sequence. The modifying caption is fed through a single layer LSTM, which represents each word as a token. MAAF simply stacks all tokens and then applies scaled dot-product attention. The output is then average pooled.}
\label{fig:model_diagram}
\end{figure*}

Our goal is to learn a common embedding space for query images modified by text and for catalog images, so that the most closely matching catalog entries for a given image + text query can be ranked by calculating the similarity between that query’s embedding and the catalog images' embeddings. Our Modality-Agnostic Attention Fusion (MAAF) approach is illustrated in Fig.~\ref{fig:model_diagram}.

The data from each modality is first passed through a standard architecture. Images are passed into a ResNet50~\cite{he2016deep,he2016identity}. Text is tokenized and word embeddings are passed through an LSTM~\cite{hochreiter1997long}. When not otherwise specified, the weights of the ResNet are pretrained on ImageNet while the word embeddings and LSTM weights are randomly initialized.

We extract token sequences from each modality and fuse them using dot-product attention. The image tokens $\phi_x$ correspond to spatial locations in ResNet feature maps. The outputs of the last convolutional block are treated as 7x7=49 tokens, one corresponding to each spatial location, after a learned linear projection to 512 dimensions from the original 2048 channels. The outputs of earlier ResNet blocks may be treated similarly; the third block yields 14x14=196 additional image tokens. Unless otherwise specified, our model uses these 49+196 = 245 image tokens for multi-resolution image understanding. The LSTM naturally yields a token $\phi_t$ for each input word token, namely the hidden state of the LSTM at that input. 

\subsection{Concatenated self-attention}

MAAF concatenates the token sequences $\Phi = [\phi_x, \phi_t]$ and passes the result into alternating self-attention and fully-connected layers as in the Transformer encoder~\cite{vaswani2017attention}. Here self-attention refers to the operation
\begin{equation}
    \text{Attn}(Q, K, V) = f(QK^T)V
\end{equation}
where each of the query $Q$, key $K$, and value $V$ matrices are calculated from the input sequence $\Phi$ by multiplying each token in the sequence by a learned matrix. Thus
\begin{equation}
    \text{Self}(\Phi) = \text{Attn}(\Phi W_Q, \Phi W_K, \Phi W_V).
\end{equation}
This mechanism allows each token to attend to any other, so an image token may be modified most strongly by another image token or by a text token. The output is another sequence of the same shape, which is added to the original sequence and possibly passed into subsequent layers with the same structure. Note that we use the same $W_Q, W_K, W_V$ for both modalities; after the initial token extraction the mechanism is agnostic to the origin of each token.

The attention map transformation $f: \R^n \rightarrow \R^n$, where $n$ is the number of key tokens (and the number of value tokens), is usually the softmax function as in \cite{vaswani2017attention}. As we discuss in sec.~\ref{sec:css}, however, we found that setting $f$ to the identity function yielded significantly better performance on the CSS task.

\subsection{Image-text cross attention}

We also experimented with treating the image tokens as one sequence and the text tokens as another, using cross-attention to allow text-image interactions as in the Transformer decoder~\cite{vaswani2017attention}. Specifically, we use cross-attention operations
\begin{equation}
    \text{Cross}(\phi_x,\phi_t) = \text{Attn}(\phi_x W_Q, \phi_t W_K, \phi_t W_V)
\end{equation}
and/or
\begin{equation}
    \text{Cross}(\phi_t,\phi_x) = \text{Attn}(\phi_t W_Q, \phi_x W_K, \phi_x W_V).
\end{equation}

MAAF allows all possible attention interactions: image-image, image-text, text-image, and text-text. Alternating self-attention within a modality and cross-attention between the modalities also allows these interactions, but we hypothesized that the modality-agnostic self attention method would work better.
We provide empirical comparisons of these methods on the Fashion IQ dataset (sec.~\ref{table:crossatt}). We note that some image-text models for other tasks, such as ViLBERT~\cite{lu2019vilbert}, have used separate image and text streams with cross-attention, while others such as VL-BERT~\cite{su2019vl} more closely resemble our modality-agnostic approach.

\subsection{Embeddings}
In order to obtain a single embedding for a given query (or catalog image), we pool the outputs across tokens. Specifically, we first average pool within each type of token, obtaining one vector of dimension 512 for the 49 image tokens corresponding to the last ResNet block, another single vector for the 196 tokens corresponding to the previous ResNet block, and another vector for the text tokens. These three vectors are then averaged, and the result is normalized. 

The pooled embedding is multiplied by a learned scale parameter to get the final embedding vector. This scalar is initialized to 4, and it is important to learn it -- leaving the scale parameter fixed instead greatly harms performance. Note this is equivalent to learning a temperature parameter in the softmax cross-entropy loss.

Catalog image embeddings are obtained in the same manner, except that there is no contribution from text tokens. Comparisons between image-text queries and catalog images are made using cosine similarity between the corresponding embeddings.

We considered several alternatives to the scheme described above for obtaining final embeddings for queries and catalog images. Table \ref{table:fiq_ablations} compares the alternatives on Fashion IQ. RP refers to our ``resolutionwise" average pooling, versus pooling all tokens equally. ITA refers to passing the catalog ``images through attention" as opposed to only through the ResNet model. IT refers to including the text tokens in the pooling, as opposed to only the tokens corresponding to image features. Table \ref{table:fiq_comparison} compares MAAF to existing methods on Fashion IQ.

\subsection{Loss and training}
We use a batch-based classification loss as in \cite{vo2019composing}, with early experiments showing that the triplet loss performs worse by the recall@k metric. Each batch is constructed from $N$ pairs of a query (image plus text) and its corresponding target image. The loss for the $i$th query is the cross-entropy between probability-like quantities
\begin{equation}
    \underset{j = 1,\ldots, N}{\text{softmax}}(\phi_{x_i t_i} \cdot \phi_{x_j})
\end{equation}
and the one-hot vector with a 1 in the $i$th position.

The model is trained at an initial learning rate of 0.001 for the image model parameters (which are pretrained on ImageNet when not otherwise stated) and 0.01 for all other parameters. The learning rate is decreased by a factor of 10 every 50k iterations and training is stopped at 150k iterations (except for the CSS dataset, see sec.~\ref{sec:css}). We use a batch size of 32 for all experiments. Our base model uses 2 attention blocks with query, key, and value sizes 512, 8 attention heads, and 256-dimensional fully-connected layers.

\section{Experiments}
We evaluate model performance on each dataset using ``recall at k" (R$k$ for short), the fraction of queries for which a labeled ``correct" result appears in the top k retrievals.
\subsection{CSS dataset}\label{sec:css}
The CSS dataset, introduced by Vo et al. in~\cite{vo2019composing}, consists of 32K queries (16K for training and 16K for test) such as ``make yellow sphere small'' that serve as modification text for synthesized images in a 3-by-3 grid scene. Although it is a relatively simple dataset, CSS has the benefit of facilitating carefully controlled experiments. We therefore use it to perform a few qualitative analyses, which we present in Sec.~\ref{sec:qualitative}, in addition to the quantitative results below.

Unlike most other datasets, larger configurations of our model trained on CSS exhibited overfitting, so we limited ourselves to configurations using block 4 (only) of ResNet50 and a single attention block. Additionally, we stopped training after 75K iterations. Replacing the softmax function with identity in the attention block significantly improved R1. Use of sinusoidal positional encodings as in \cite{vaswani2017attention} did not significantly improve performance when using softmax, and significantly impaired performance when using the identity function.

\begin{table}
\begin{center}
\begin{tabular}
{|p{6cm}|c|}
\hline
Method & R1 \\
\hline
\hline
Show and Tell~\cite{vinyals2015show} & 33.0 $\pm$ 3.2 \\
Parameter hashing~\cite{noh2016image} & 60.5 $\pm 1.9$ \\
Relationship~\cite{santoro2017simple} & 62.1 $\pm1.2$ \\
FiLM~\cite{perez2018film} & 65.6 $\pm$ 0.5 \\
TIRG~\cite{vo2019composing} & 73.7 $\pm$ 1.0 \\
Locally Bounded Features~\cite{Hosseinzadeh_2020_CVPR} & 79.2 $\pm$ 1.2 \\
MAAF (ours) & \textbf{87.8 $\pm$ 0.9} \\
\hline
\end{tabular}
\end{center}
\caption{Results on the CSS test set, with comparisons to existing methods as reported in~\cite{vo2019composing, Hosseinzadeh_2020_CVPR}, and our reproduction of TIRG based on code provided by the authors. \label{table:css}}
\end{table}

\subsection{Fashion IQ}\label{sec:fiq}
Fashion IQ~\cite{guo2019fashion} consists of 
$\sim$60K real-world product images (originally sourced from {\it amazon.com}) with human natural language sentences to facilitate the training of interactive image retrieval systems. The sentences are derived by tasking annotators with describing the differences between each target product image and a reference product image (e.g., ``longer more dressy''). The product items were selected from three categories: Dresses, Tops\&Tees, and Shirts. The training data consists of $\sim$36K examples, with $\sim$12K examples each for the validation and test splits. Within each split, examples are approximately equally divided across the three categories. 
The strongest models in competitions at recent conference workshops\footnote{ https://sites.google.com/view/lingir/fashion-iq} were ensembles of weaker single models. In this paper we focus on single (not ensembled) models. 

Since the three categories are fairly distinct, we draw each minibatch from a single category.

\subsubsection{Comparisons to reported results and ablations}
We find that our best model outperforms all other reported single models on the validation set, with the exception of a model reported in a technical report \cite{li2019designovel} that used extra non-public data. Since the model of \cite{li2019designovel} is very similar to TIRG~\cite{vo2019composing} and results for their model trained without this extra non-public data are not available, we speculate that this text corpus is important to their increased performance relative to TIRG. We compare also to TIRG directly and find that our model performs better.
See Tables~\ref{table:fiq_ablations} and ~\ref{table:fiq_comparison} for full results, including detailed ablations of the design choices for our attention mechanism.

Other competition entries outperformed our best single model using ensembles of weaker models~\cite{yu2019curlingnet,zhao2019superraptors}.

\begin{table*}
\begin{center}
\begin{tabular}
{|l|c|c|c|c|c|c|c|c|c|c|c|}
\hline
Model & RP & ITA & IT & Dresses R10 (R50) & Tops\&Tees R10 (R50) & Shirts R10 (R50) & $\frac{\text{R10 + R50}}{2}$ \\
\hline
\hline
 Ours & - & - & - & 21.6 $\pm$ 0.6 (44.9 $\pm$ 0.6) & 25.5 $\pm$ 0.4 (50.6 $\pm$ 0.5) & 19.0 $\pm$ 0.6 (40.6 $\pm$ 0.7) & 33.7 $\pm$ 0.3 \\
 Ours & - & - & \checkmark & 20.9 $\pm$ 1.0 (44.1 $\pm$ 0.5) & 26.2 $\pm$ 0.8 (51.0 $\pm$ 0.8) & 19.1 $\pm$ 0.8 (40.9 $\pm$ 1.1) & 33.7 $\pm$ 0.2 \\
 Ours & - & \checkmark & - & 24.0 $\pm$ 1.1 (48.1 $\pm$ 0.9) & 27.1 $\pm$ 0.7 (53.1 $\pm$ 1.5) & 20.5 $\pm$ 0.3 (42.2 $\pm$ 0.2) & 35.8 $\pm$ 0.5 \\
 Ours & - & \checkmark & \checkmark & 23.9 $\pm$ 0.6 (48.5 $\pm$ 0.2) & 27.6 $\pm$ 0.4 (53.5 $\pm$ 0.8) & 21.0 $\pm$ 0.8 (43.8 $\pm$ 0.5) & 36.4 $\pm$ 0.2 \\
 Ours & \checkmark & - & - & 21.6 $\pm$ 0.9 (45.6 $\pm$ 1.7) & 25.4 $\pm$ 0.8 (50.7 $\pm$ 1.1) & 19.4 $\pm$ 0.8 (40.2 $\pm$ 1.1) & 33.8 $\pm$ 0.8 \\
 Ours & \checkmark & - & \checkmark & 21.5 $\pm$ 0.6 (45.5 $\pm$ 0.6) & 25.5 $\pm$ 0.9 (51.4 $\pm$ 0.9) & 19.4 $\pm$ 0.7 (41.3 $\pm$ 0.9) & 34.1 $\pm$ 0.3 \\
 Ours & \checkmark & \checkmark & - & 24.1 $\pm$ 0.5 (48.2 $\pm$ 0.8) & 27.7 $\pm$ 0.7 (52.5 $\pm$ 0.5) & 20.8 $\pm$ 0.5 (43.2 $\pm$ 0.9) & 36.1 $\pm$ 0.4 \\
 Ours &  \textbf{\checkmark} & \textbf{\checkmark} & \textbf{\checkmark} & \textbf{23.8 $\pm$ 0.6 (48.6 $\pm$ 1.0)} & \textbf{27.9 $\pm$ 0.8 (53.6 $\pm$ 0.6)} & \textbf{21.3 $\pm$ 0.7 (44.2 $\pm$ 0.3)} & \textbf{36.6 $\pm$ 0.4} \\
\hline
\end{tabular}
\end{center}
\caption{Results on the Fashion IQ validation set, with ablations and comparison to other existing methods. All rows used 7x7 and 14x14 inputs with 2 attention blocks and 256 unit fully connected layers. RP = `Resolution-wise pooling', ITA = `Image through attention', IT = `Include text'. * indicates a model that used non-public extra data for training. Bolding indicates the best result using only publicly-available data. See Section \ref{sec:ablations} for more details. \label{table:fiq_ablations}}
\end{table*}

\begin{table}
\begin{center}
\begin{tabular}
{|l|c|c|}
\hline
Model & AD & $\frac{\text{R10 + R50}}{2}$ \\
\hline
\hline
TIRG~\cite{vo2019composing} & & 31.20 \\
VAL~\cite{Chen_2020_CVPR} & & 35.4 \\
MAAF (Ours) & & \textbf{36.6 $\pm$ 0.4} \\
Designovel~\cite{li2019designovel} & \textbf{\checkmark}  & 37.28 \\
Designovel+augmentation & \textbf{\checkmark} & 40.84 \\

\hline
\end{tabular}
\end{center}
\caption{Results on the Fashion IQ validation set, with comparison to other existing methods. AD indicates a model that used non-public extra data for training. Bolding indicates the best reported result for a single model using only publicly-available data. \label{table:fiq_comparison}}
\end{table}

\subsubsection{Attention model variations}\label{sec:crossatt}
We also investigated several two-stream variations of our attention mechanism to confirm that our primary single-stream variant performs best. Results of those experiments are shown in Table \ref{table:crossatt}.

\begin{table*}
\begin{center}
\begin{tabular}
{|l|c|}
\hline
Method & $\frac{\text{R10 + R50}}{2}$ \\
\hline
\hline
\text{Cross}($\phi_x,\phi_t$) & 32.4 $\pm$ 0.2 \\ %
\text{Self}($\phi_x$),\  \text{Cross}($\phi_x,\phi_t)$ & 34.4 $\pm$ 0.4 \\ %
\text{Cross}$(\phi_x,\phi_t)$ $||$ \text{Cross}$(\phi_t,\phi_x)$ & 33.2 $\pm$ 0.1 \\ %
\text{Self}$(\phi_x)$ $||$ \text{Self}$(\phi_t)$, \ \text{Cross}$(\phi_x,\phi_t)$,\ \text{Cross}$(\phi_t,\phi_x)$ & 32.7 $\pm$ 0.3 \\ %
\text{Self}$(\phi_x)$ $||$ \text{Self}$(\phi_t)$, \  \text{Cross}$(\phi_t,\phi_x)$,\ \text{Cross}$(\phi_x,\phi_t)$ & 30.4 $\pm$ 0.2 \\ %
\text{Self}$(\phi_x)$ $||$ \text{Self}$(\phi_t)$, \  \text{Cross}$(\phi_x,\phi_t)$ $||$ \text{Cross}$(\phi_t,\phi_x)$ & 33.5 $\pm$ 0.3 \\
MAAF: \text{Self}$(\Phi)$ & \textbf{36.6 $\pm$ 0.4} \\
\hline
\end{tabular}
\end{center}
\caption{Fashion IQ validation scores (mean $\frac{\text{R10 + R50}}{2}$ over categories) for several architecture variations, where $||$ denotes parallel branches, and sequential layers are separated by a comma. Average and standard deviation are computed over 5 trials with different random initializations.   \label{table:crossatt}}
\end{table*}

\subsubsection{Text Representation}
Self-supervised training of the Transformer model on large text corpora has proved effective for a variety of natural language tasks, including natural language inference, sentiment analysis, named entity recognition, and question answering~\cite{devlin2018bert}. A natural approach to encode a text sentence is to use this type of model to extract its highly contextual word feature representations.
However, for Fashion IQ, which has only 4,769 words in the training set vocabulary, and whose modifying sentences contain only 5.4 tokens on average, it is not clear that such a large neural network is necessary or optimal for the problem. Therefore we did a comparative study of different sentence encodings.

Table \ref{table:text_model} compares the performance of MAAF when using only word embeddings (Embedding), versus the sequential word representations from a recurrent neural network (LSTM), versus context-aware word representations extracted by a self-attention model, learnable (Transformer) or pretrained (BERT). The experiments revealed that learning an LSTM from randomly initialized word embeddings is most effective for this task. A single embedding layer that treats the sentence as a bag of words performs only about 27\% (relative) worse than the LSTM. We attribute this result to the simplistic nature of the modifying text sentences. Somewhat surprisingly, richer word context does not seem to benefit the task as evidenced by the weaker score obtained by a transformer with a single encoder block. 
Additionally, the model with large pretrained BERT for the text model achieves higher validation performance when the BERT weights are frozen during training.

Fixing the embedding layer with the GLOVE weight matrix seems to hurt performance, with the exception of a single linear layer, where fixed GLOVE word embeddings impose a useful prior on the word distribution. We attribute this to the small size and fashion-specific domain of the vocabulary, which a co-occurrence matrix for more general vocabulary maps to a small region of the embedding space.     

\begin{table*}
\begin{center}
\begin{tabular}
{|l|c|c|c|c|c|}
\hline
Method & Dresses R10(R50) & Tops\&Tees R10(R50) & Shirts R10(R50) & Mean R10(R50) & $\frac{\text{R10 + R50}}{2}$ \\
\hline
\hline
Embedding Layer & 18.84 (39.76) & 21.72 (45.64) & 16.49 (36.31) & 19.02 (40.57) & 29.79 \\
Linear Layer with GLOVE & 21.42 (43.23) & 24.48 (50.13) & 17.81 (40.19) & 21.24 (44.52) & 32.88 \\
Linear Layer with GLOVE* & 22.36 (47.00) & 27.13 (50.79) & 21.10 (41.46) & 23.53 (46.42) & 34.97 \\
\hline
LSTM & \textbf{26.03 (50.62)} & \textbf{28.25 (55.38)} & 20.12 (43.13) & \textbf{24.80 (49.71)} & \textbf{37.25} \\
LSTM with GLOVE & 24.24 (46.75) & 28.05 (54.56) & \textbf{21.30 (44.11)} & 24.53 (48.48) & 36.50 \\
\hline
Transformer & 21.32 (44.17) & 25.55 (50.94) & 19.19 (40.14) & 22.02 (45.09) & 33.55 \\
Transformer with GLOVE & 20.77 (43.23) & 24.78 (48.95) & 18.45 (38.86) & 21.34 (43.68) & 32.51 \\
\hline
BERT & 18.59 (39.66) & 23.05 (45.95) & 18.55 (37.63) & 20.06 (41.08) & 30.57 \\
BERT (frozen) & 22.61 (43.23) & 25.45 (49.26) & 19.68 (41.61) & 22.58 (44.70) & 33.64 \\
\hline
\end{tabular}
\end{center}
\caption{Impact of different text model architectures on Fashion IQ Retrieval results. All model weights with the exception of BERT and GLOVE weights are randomly initialized. We also present the benefit of using GLOVE~\cite{pennington2014glove} frozen pretrained word embeddings on each of the models. In all our experiments, the learning of the image model is set to 1/10 of the current learning rate, whereas the text pathway is updated with the ordinary learning rate. Using the smaller learning rate for the text model weights only improved performance for Linear Layer with GLOVE; this experiment is marked with an asterisk. \label{table:text_model}}
\end{table*}

\subsection{Birds-to-Words}
Birds-to-Words~\cite{forbes2019neural} consists of images of birds from iNaturalist together with paragraphs humans wrote to describe the difference between pairs of these images. Each of 3,347 image pairs has on average 4.8 paragraphs, each describing the differences between the pair of birds in an average of 32.1 tokens. Birds-to-Words provides richer text descriptions in each example than any of the other datasets we study, although the number of examples is small. Forbes et al.~\cite{forbes2019neural} studied the generation of these fine-grained relative captions, but we adapt their dataset for the task of retrieving the described second image given the relative caption. We train our model according to the best configuration described in Section \ref{sec:fiq}, but for only 30k batches to avoid overfitting. The baseline models are trained with the same parameters for the shared components, with the number of training iterations selected to optimize validation set performance. Table \ref{table:birdsandspot} shows our results for this task on the Birds-to-Words test set, which does not share images or descriptions with the training or validation sets.

\begin{table}
\begin{center}
\begin{tabular}
{|l|c|c|c|c|}
\hline
 & \multicolumn{2}{|l|}{Birds-to-Words} & \multicolumn{2}{|l|}{Spot-the-Diff}\\
Method & R10 & R50 & R10 & R50 \\
\hline
\hline
Image only & 21.8 & 51.0 & 7.05 & 22.33  \\
Text only & 20.0 & 49.4 & 2.00 & 6.67  \\
TIRG & 27.7 & 51.3 & 8.19 & 25.10  \\
MAAF (Ours) & \textbf{34.75} & \textbf{66.29} & \textbf{10.14} & \textbf{26.38}  \\
\hline
\end{tabular}
\end{center}
\caption{Results on the Birds-to-Words and Spot-the-Diff test queries, with comparisons to baselines. \label{table:birdsandspot}}
\end{table}

\subsection{Spot-the-Diff}

Spot-the-Diff~\cite{jhamtani2018learning} consists of snapshots from surveillance footage together with single-sentence captions humans wrote to describe the differences between pairs of snapshots from one of 11 viewpoints at different times. There are 17,672 image-image-caption triples from 8,566 image pairs in the training set, 3,310 triples from  1,493 image pairs in the validation set, and 2,107 triples from 1,270 image pairs in the test set. The dataset was originally designed to study relative caption generation, but it provides a challenging setting for our task since images from the same source are highly similar. Table \ref{table:birdsandspot} shows results for our model and baselines on the test set.

\subsection{Fashion200k}
The Fashion200k dataset~\cite{han2017automatic} as used in Vo et al.~\cite{vo2019composing} consists of pairs of images of women's clothing products that differ by one attribute. The word for that attribute serves as the modifying caption.
We took the best performing configuration of our model on Fashion IQ, and trained it in the same way on Fashion200k with essentially no changes. Table \ref{table:fashion200k} compares the result to existing methods.

\begin{table}
\begin{center}
\begin{tabular}
{|l|c|}
\hline
Method & R1 \\
\hline
\hline
Han et al. \cite{han2017automatic} & 6.3 \\
Show \& Tell~\cite{vinyals2015show} & 12.3 $\pm$ 1.1  \\
FiLM~\cite{perez2018film} & 12.9 $\pm$ 0.7  \\
Param hashing~\cite{noh2016image} & 12.2 $\pm 1.1$ \\
Relationship\cite{santoro2017simple} & 13.0 $\pm$ 0.6 \\
TIRG~\cite{vo2019composing} & 14.1 $\pm$ 0.6 \\
Locally Bounded~\cite{Hosseinzadeh_2020_CVPR} & 17.8 $\pm$ 0.5 \\
VAL~\cite{Chen_2020_CVPR} & \textbf{22.9} \\
Ours & 18.94 \\
\hline
\end{tabular}
\end{center}
\caption{Results on the Fashion 200K test queries, with comparisons to existing methods and our reproduction of TIRG based on code provided by the authors. \label{table:fashion200k}}
\end{table}

\section{Qualitative results and visualizations}\label{sec:qualitative}
The simplicity of the CSS dataset (see sec.~\ref{sec:css}) allows us to probe our attention mechanism and visualize its behavior. We analyzed our model with softmax on this dataset, using only the 7x7 representation from block 4 of ResNet, and only one attention block. While each attention head may do something different, we observed systematic behavior by averaging attention maps over the attention heads and over data examples with particular properties.

\begin{figure}
\begin{center}
\includegraphics[width=0.75\linewidth]{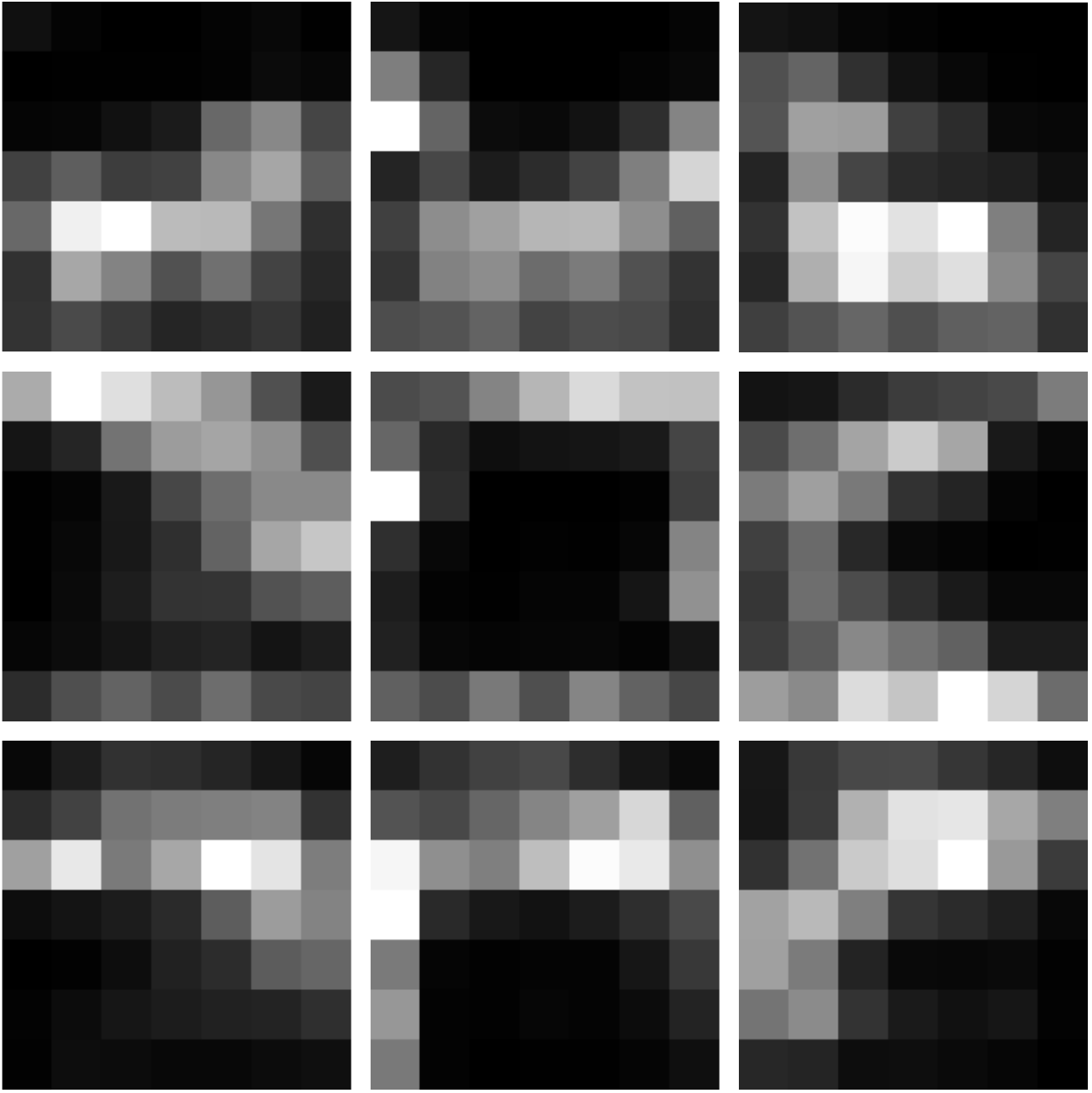}
\end{center}
   \caption{Averaged 7x7 attention maps corresponding to location word tokens from the CSS test set (averaged across all eight attention heads): top-left, top-center, top-right, middle-left, middle-center, middle-right, bottom-left, bottom-center, and bottom-right. White indicates larger attention weights.}
\label{fig:css_position}
\end{figure}

\begin{figure}
\begin{center}
\includegraphics[width=0.9\linewidth]{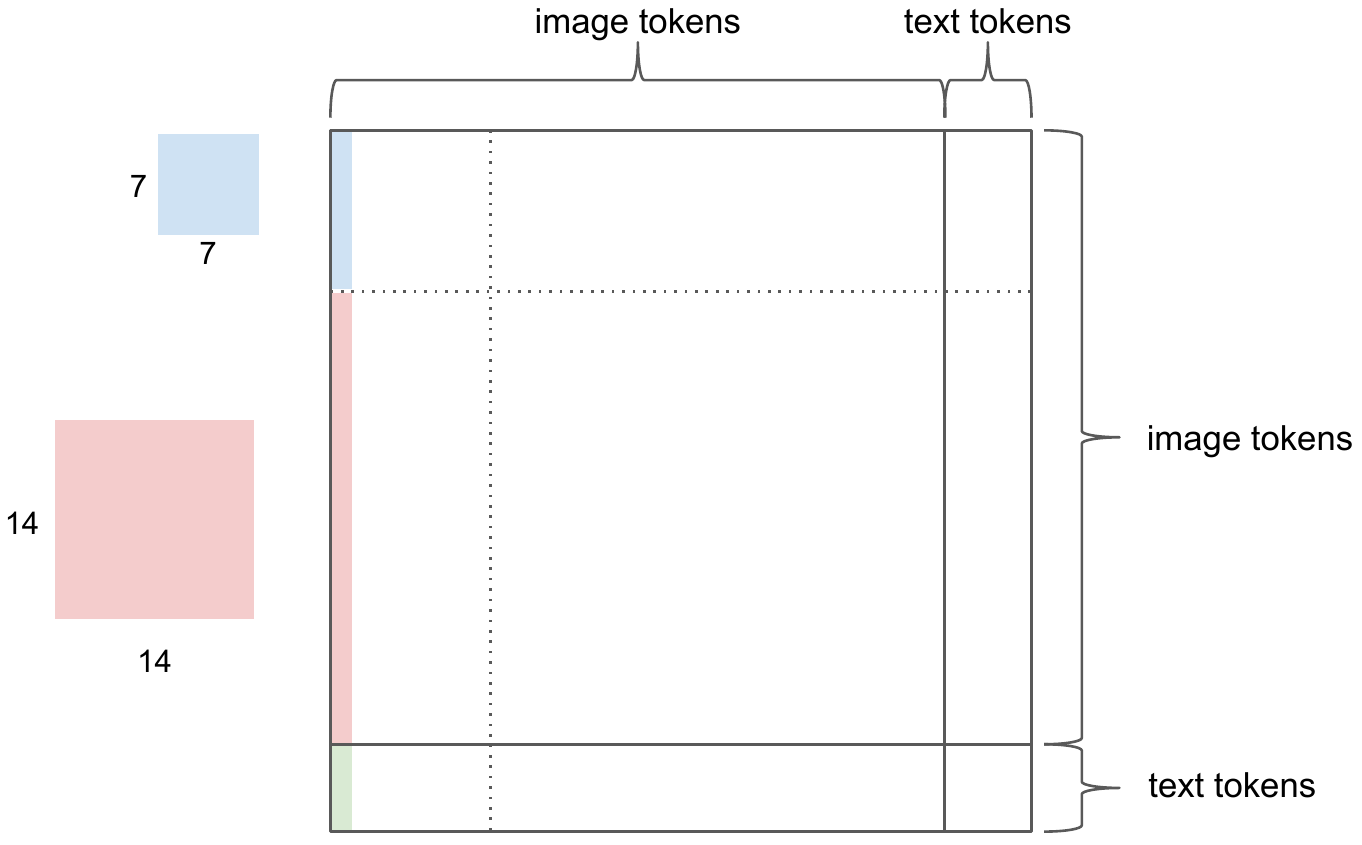}
\end{center}
   \caption{The structure of $f(QK^T)$ in concatenated self-attention, annotated with approximate areas devoted to each `type' of attention (image to image vs. image to text). The blue square represents the 7x7 array of image tokens extracted from block 4 of ResNet; the red square represents the 14x14 array of image tokens extracted from block 3. Each 2d array of tokens is reshaped to a 1d array of tokens by stacking. Finally, the text tokens, shown in green, are added to the bottom. }
\label{fig:attentionstructure}
\end{figure}

For Fig.~\ref{fig:css_position}, we computed the attention weights $f(QK^T)$ in our trained model for all text+image examples in the CSS test set, then averaged them across attention heads. We then divided the examples into groups by position word (e.g. ``top-right", ``bottom-center'', etc.), averaged over these groups, extracted the first 49 elements of the row of $f(QK^T)$ corresponding to the token for the position word, and reshaped these elements into a 7x7 array. Fig.~\ref{fig:attentionstructure} illustrates the structure of $f(QK^T)$ to help the reader understand this procedure. We thus obtained an image of how strongly, on average, ``top-right" attends to each of the 7x7 image regions. Notice that in all cases, the region referenced by the position word is a dark spot surrounded by higher weights. The position word does not attend to the position it references; rather it attends to \textit{everything else}.

This behavior -- a word systematically ignoring what it references -- seems to contradict the intuition provided by image captioning work. For example, Xu et al. \cite{xu2015show} show that, in an attention mechanism linking words and text, words will in general attend to relevant regions of an image. Because in our setting the captions describe modifications to the image, and the model seeks a final embedding closer to the target image, a word may avoid attending to its referent since its referent in the query image will tend not to be informative about the target image.

\begin{figure}
\begin{center}
\includegraphics[width=\linewidth]{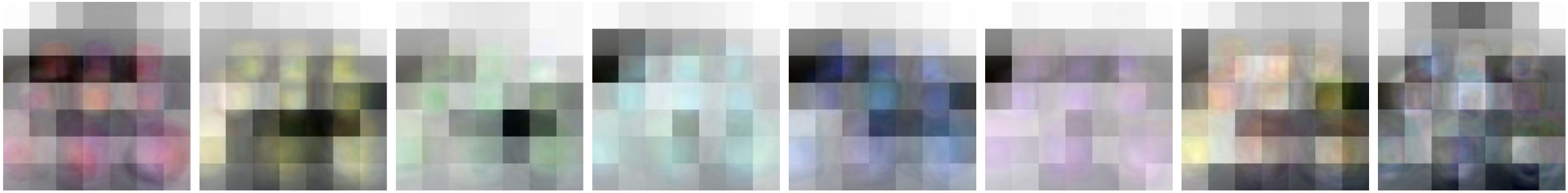}
\end{center}
   \caption{Averaged candidate images from the CSS test set, multiplicatively modulated by one minus the 7x7 attention map (averaged across all eight attention heads) produced by our network with a single attention block for color word tokens: red, yellow, green, cyan, blue, purple, brown and gray.}
\label{fig:css_color}
\end{figure}

Fig.~\ref{fig:css_color} shows that this counter-intuitive pattern appears to hold for color words as well. Here we use not the attention weights for the color word directly, but one minus these weights. The result is, for example, an image of what the word ``red'' \textit{ignores} in the image, that is, what the word ``red'' systematically does \textit{not} pay attention to. Fig.~\ref{fig:css_color} shows the 1-minus-attention-modulated images averaged over examples containing each color word (excluding examples where the referenced object is being added), with the overall mean image subtracted. We again see that each color word systematically \textit{ignores} the object to which it refers (i.e., an object of that color).

\begin{figure}
\begin{center}
\includegraphics[width=0.99\linewidth]{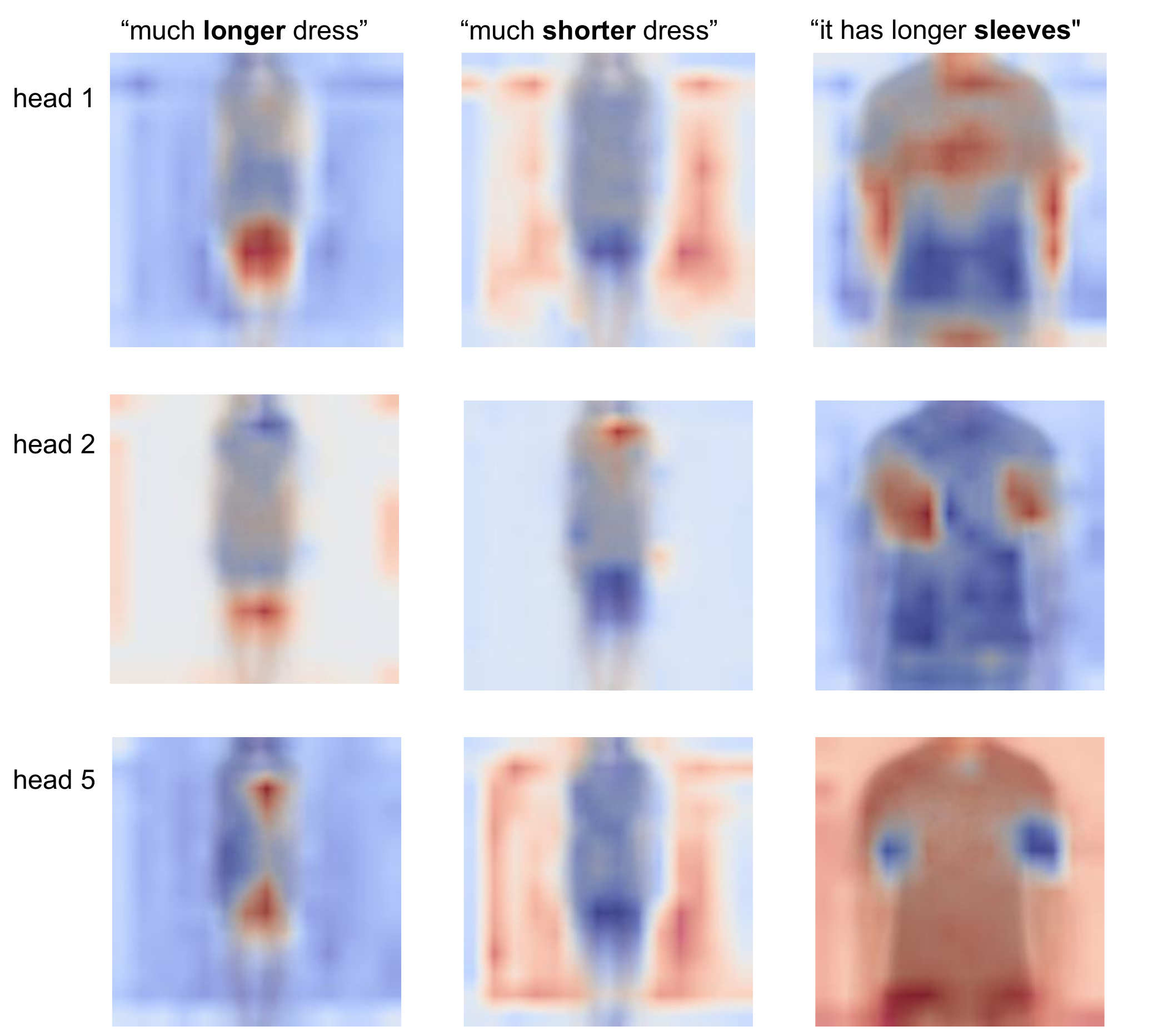}
\end{center}
   \caption{Attention maps for the feature representation of the bold word in the sentence, with respect to the positional image feature vectors. Attention maps are represented for three different attention heads (among the 8 in a block of our best performing model), and averaged over the first 200 images of the test split of the FashionIQ dataset.}
\label{fig:fashioniq_attention}
\end{figure}

Figure \ref{fig:fashioniq_attention} shows analogous images for Fashion IQ, showing some of the related effects that can occur in a less controlled setting than CSS. For words in the dataset that correspond to specific parts of a clothing item, such as short sleeves, or longer/shorter dress, we notice that several attention heads highlight the corresponding parts of the image. More interestingly, for additive changes such as ``longer dress", the model shows high attention weights over the corresponding image region, whereas for subtractive changes such as "shorter dress", the model has negative attention weights over the corresponding image region, focusing on the remaining parts to keep.

\section{Related Work}
The ability of modern deep neural network architectures to extract highly semantic image features, as well as contextual and long-range text features, has ignited the study of tasks that combine image and natural language understanding. Most of the work in this field has been spurred by initial efforts to generate more discursive image caption descriptions (Image Captioning  \cite{vinyals2015show,show2015tell,rennie2017self,anderson2018bottom,sharma2018conceptual}), work often generalized to visual question answering (VQA) \cite{antol2015vqa,lu2016hierarchical,fukui2016multimodal,anderson2018bottom}. Other tasks such as visual grounding of text (so-called referring expressions), visual commonsense reasoning, and caption-based image retrieval have also received significant attention.

\subsection{Fine-grained image retrieval}
Several image retrieval challenges and corresponding datasets have fostered the study of domain-specific image feature extraction. In the fashion domain, datasets like Street-to-Shop~\cite{hadi2015buy}, DeepFashion~\cite{liu2016deepfashion}, and DeepFashion-v2~\cite{ge2019deepfashion2} are labeled with clothing item attributes, as well as which images correspond to the same product. The triplet loss was introduced as a way to learn from these paired label annotations in~\cite{hoffer2015deep}, and several sampling techniques for mining hard negatives were studied in subsequent works ~\cite{schroff2015facenet,hermans2017defense}. Other work explored variations of the loss function and related tasks such as face identification ~\cite{hoffer2015deep,deng2019arcface,wang2018cosface}.

\subsection{Image and text modeling with attention}
Hu et al. in~\cite{hu2018relation} demonstrate the benefit of incorporating attention for object recognition. The authors propose a ``geometric attention'' that seeks to account for the spatial relationships between object pairs. On the text side, several works have shown the importance of attention modeling for improved performance in tasks such as machine translation~\cite{bahdanau2014neural,vaswani2017attention} and language modeling~\cite{devlin2018bert,dai2019transformer}. Most notably, the introduction of the Transformer architecture in~\cite{vaswani2017attention}, with its ability to model both self- and cross-attention, has enabled significant advances in the state-of-the-art for these and other tasks. Self-supervised pretraining on a large text corpus allowed BERT~\cite{devlin2018bert} to achieve still better results.

\subsection{Joint visuolinguistic representations}
Anderson et al. used text context to weight attention to image regions proposed by a bottom-up object detector for image captioning and VQA~\cite{anderson2018bottom}. Previous approaches used spatial locations in a CNN feature map as objects of attention for these tasks ~\cite{rennie2017self,lu2017knowing,yang2016review,xu2015show,lu2016hierarchical,xu2016ask,yang2016stacked,zhu2016visual7w}.

Parallel to our work, several groups have developed models using attention mechanisms for joint visuolinguistic representations for tasks such as visual commonsense reasoning, visual question answering, referring expression comprehension, and image retrieval from captions. Several recent works~\cite{su2019vl,tan2019lxmert,li2019visualbert,lu2019vilbert,zhou2019unified,alberti2019fusion,chen2019uniter} train BERT-like models on large-scale data to achieve general visuolinguistic representations which are then fine-tuned (possibly with modifications or additions) for these tasks. A similar model has also been trained directly on image captioning and VQA~\cite{liu2019aligning}. These models all extract features from the region proposals of an object detector, which is more appropriate for understanding scenes composed of multiple familiar objects than the fine-grained differences in single objects that we address. Another approach for related problems uses image patches~\cite{gao2020fashionbert}.
We instead use the positional feature maps at multiple scales as image tokens, to learn both global (e.g. shape, size) and local features (e.g. texture, material) which an object detector cannot decompose.

Hosseinzadeh and Wang~\cite{Hosseinzadeh_2020_CVPR} approached the same task we address using fixed regions of image pixels to obtain image tokens for a cross-attention fusion mechanism similar to the ones we studied in section \ref{sec:crossatt}. Their approach is weaker on the datasets where comparison is possible. 

Chen, Gong, and Bazzani~\cite{Chen_2020_CVPR} use outputs at multiple levels of a convolutional network to obtain image features at multiple scales for attention as we do, but their model fuses the text as a single vector with the image features before their attention mechanisms. This approach is substantially different and does not perform as well as ours on Fashion IQ, although it performs better on Fashion 200k, which uses single-word text modifications. We suspect that their approach is better suited for simple text whereas ours is better adapted for longer, more complicated text modifications.

{\small
\bibliographystyle{ieee_fullname}
\bibliography{main}
}

\beginsupplement
\section{Detailed model ablations on Fashion IQ}\label{sec:ablations}
Table \ref{table:fiq_ablations} shows the performance of several variations of the attention fusion mechanism on the Fashion IQ validation set, including means and standard deviations across five runs with identical settings but different random initializations. Below we clarify the meaning of each variation of our model.

A model with resolution-wise pooling (RP) performs average-pooling across the 7x7=49 image tokens from the last ResNet block separately from average-pooling across the 14x14 image tokens from the penultimate ResNet block. The results of pooling each of the two sets of tokens are then averaged to obtain the final embedding. A model \textit{without} resolution-wise pooling simply treats all image tokens equally. We hypothesized that resolution-wise pooling would provide some benefit by correcting the over-representation of the finer-grained 14x14 features. Our experiments supported that hypothesis, but the size of the performance increase was comparable to the inter-trial variation.

A model with ``image through attention" (ITA) uses the attention blocks even in the absence of a modifying caption. ITA can also be thought of as modifying the catalog images by the ``null" caption containing no words, using the MAAF mechanism. A model without ITA directly uses ResNet features to represent the catalog images. Models with ITA consistently perform better by around 2 percentage points, suggesting that this is a better formulation for the task and an important part of our model. Note that ITA does not imply additional parameters, since the same attention blocks are used for catalog images as for the image-plus-text queries.

A model with ``include text" (IT) includes the outputs corresponding to the text tokens in the final average pooling, allowing these tokens to directly influence the final embedding. Using this variation tends to have a small benefit.

Table~\ref{table:fiq_sweep_att_layer_spec} shows the effect of ablating the 14x14 resolution inputs, while keeping RP, IT, and ITA (RP has no effect when only 7x7 inputs are utilized). We also tested the sensitivity of our model to the number of units in the fully connected layers following the dot product attention layers (Table~\ref{table:fiq_sweep_width}) and the number of transformer blocks in our network (Table~\ref{table:fiq_sweep_number_attention_blocks}).

Table \ref{table:fiq_sweep_width} shows the performance of MAAF on Fashion IQ as we vary the size of the fully-connected layers in the attention blocks. The result is not very sensitive to this hyperparameter, and all other results in the paper use our original setting of 256. 

\begin{table*}
\begin{center}
\begin{tabular}
{|c|c|c|c|c|c|c|c|c|c|c|c|c|}
\hline
7x7  & 14x14  & Dresses R10 (R50) & Tops\&Tees R10 (R50) & Shirts R10 (R50) & $\frac{\text{R10 + R50}}{2}$ \\
\hline
\hline
\checkmark & - & 22.9 $\pm$ 1.1 (47.7 $\pm$ 1.0) & 26.4 $\pm$ 0.6 (52.4 $\pm$ 0.5) & 19.5 $\pm$ 0.9 (42.0 $\pm$ 0.6) & 35.1 $\pm$ 0.6 \\
\checkmark & \checkmark & \textbf{23.8 $\pm$ 0.6 (48.6 $\pm$ 1.0)} & \textbf{27.9 $\pm$ 0.8 (53.6 $\pm$ 0.6) } & \textbf{21.3 $\pm$ 0.7 (44.2 $\pm$ 0.3)} & \textbf{36.6 $\pm$ 0.4} \\
\hline
\end{tabular}
\end{center}
\caption{Results on the Fashion IQ validation set, with and without the higher spatial resolution input (14x14). All rows used two transformer blocks with 256 unit fully connected layers, RP, ITA, and IT. The second row is our standard MAAF model.\label{table:fiq_sweep_att_layer_spec}}
\end{table*}

\begin{table*}
\begin{center}
\begin{tabular}
{|l|c|c|c|c|c|c|c|c|c|c|c|}
\hline
Width & Dresses R10 (R50) & Tops\&Tees R10 (R50) & Shirts R10 (R50) & $\frac{\text{R10 + R50}}{2}$ \\
\hline
\hline
64 & 23.9 $\pm$ 0.8 (48.6 $\pm$ 0.8) & 27.7 $\pm$ 1.0 (53.7 $\pm$ 0.8) & 21.1 $\pm$ 0.5 (43.3 $\pm$ 0.6) & 36.4 $\pm$ 0.3 \\
128 & \textbf{ 24.7 $\pm$ 0.6 (49.0 $\pm$ 0.7)} & 27.8 $\pm$ 0.6 (\textbf{54.1 $\pm$ 0.3}) & 21.1 $\pm$ 0.4 (43.9 $\pm$ 0.9) & \textbf{36.8 $\pm$ 0.2} \\
256 & 23.8 $\pm$ 0.6 (48.6 $\pm$ 1.0) & \textbf{27.9 $\pm$ 0.8} (53.6 $\pm$ 0.6) & \textbf{21.3 $\pm$ 0.7 (44.2 $\pm$ 0.3)} & 36.6 $\pm$ 0.4 \\
512 & 24.2 $\pm$ 0.4 (48.3 $\pm$ 0.4) & 27.7 $\pm$ 0.8 (53.8 $\pm$ 1.1) & 21.2 $\pm$ 1.1 (43.2 $\pm$ 0.3) & 36.4 $\pm$ 0.5 \\
\hline
\end{tabular}
\end{center}
\caption{Results on the Fashion IQ validation set, tuning the number of model parameters by varying the width of the fully connected layer of the attention blocks. All rows used two attention blocks with 7x7 and 14x14 inputs, RP, ITA, and IT.\label{table:fiq_sweep_width}}
\end{table*}

\begin{table*}
\begin{center}
\begin{tabular}
{|l|c|c|c|c|c|c|c|c|c|c|c|}
\hline
\# Attention Blocks & Dresses R10 (R50) & Tops\&Tees R10 (R50) & Shirts R10 (R50) & $\frac{\text{R10 + R50}}{2}$ \\
\hline
\hline
1 & \textbf{24.2 $\pm$ 1.2} (48.3 $\pm$ 1.3) & \textbf{27.9 $\pm$ 0.8 (54.0 $\pm$ 0.9)} & 20.6 $\pm$ 0.7 (43.3 $\pm$ 0.6) & 36.4 $\pm$ 0.5 \\
2 & 23.8 $\pm$ 0.6 (\textbf{48.6 $\pm$ 1.0}) & 27.9 $\pm$ 0.8 (53.6 $\pm$ 0.6) & \textbf{21.3 $\pm$ 0.7 (44.2 $\pm$ 0.3)} & \textbf{36.6 $\pm$ 0.4} \\
3 & 24.1 $\pm$ 0.3 (48.3 $\pm$ 0.4) & 27.3 $\pm$ 0.7 (53.6 $\pm$ 0.9) & 20.6 $\pm$ 0.2 (43.5 $\pm$ 0.7) & 36.2 $\pm$ 0.4 \\
4 & 23.7 $\pm$ 0.8 (47.9 $\pm$ 1.0) & 27.1 $\pm$ 0.8 (52.9 $\pm$ 1.0) & 20.0 $\pm$ 0.5 (42.6 $\pm$ 0.7) & 35.7 $\pm$ 0.6 \\
\hline
\end{tabular}
\end{center}
\caption{Results on the Fashion IQ validation set, tuning the number of attention blocks. All rows used 256 unit fully connected layers with 7x7 and 14x14 inputs, RP, ITA, and IT.\label{table:fiq_sweep_number_attention_blocks}}
\end{table*}

\subsection{Cross-attention variations}
MAAF uses perhaps the simplest dot-product attention mechanism for fusing representations from different modalities: we concatenate the ``sequences" of tokens from each modality and treat each token in the resulting sequence equally. We also studied two-stream approaches where the architecture explicitly keeps the two modalities separate. To fairly compare the two-stream approach to our one-stream approach, we evaluated several variations of the two-stream approach in Table \ref{table:crossatt} of the main text. Here we clarify the computation performed by each of these models and why we expected that the one-stream approach would perform better.

Table \ref{table:crossatt} shows the performance of several combinations of attention operations on Fashion IQ. In each case, a sequence of attention operations (each treated as a residual) is followed by a position-wise feedforward layer. Commas indicate operations performed in sequence, parallel bars $||$ indicate operations done in parallel, \textit{i.e.}, the second operation does not use the outputs of the first operation.

We found that a model including all four operations in sequence did not train stably under our training scheme. We therefore used alternative learning rate schedules: dividing all rates by 2, and a simple ``warm-up" starting with this lower rate for 5000 iterations. Results in Table \ref{table:crossatt} are reported for the best learning rate schedule for each model, usually the largest rate leading to stable training.

To see in detail how the different attention combinations relate to one another and to the single-stream approach, let $x^0$ denote the ``sequence" of image tokens as input to the attention block and $y^{0}$ denote the sequence of text tokens. Then one head of image self-attention with a residual connection is (suppressing LayerNorm and Dropout for brevity)
\begin{equation}
    x^{\alpha+1} = x^{\alpha} + f( x^{\alpha}W_Q W_K^T x^{\alpha} ) x^{\alpha}W_V 
\end{equation}
and cross-attention to the text is
\begin{equation}
    x^{\alpha+1} = x^{\alpha} + f( x^{\alpha}W_Q W_K^T y^{\alpha} ) y^{\alpha}W_V .
\end{equation}
The most competitive two-stream attention block we found acts on image tokens in the following way
\begin{align} \nonumber
    x^{1} &=  x^{0} + f( x^{0}W_{Q0} W_{K0} x^{0} ) x^{0}W_{V0}  \\ \nonumber
    \Rightarrow x^{2} &= x^{1} + f( x^{1}W_{Q1} W_{K1} y^{0} ) y^{0}W_{V1}  \\ \nonumber
    &= x^{0} + f( x^{0}W_{Q0} W_{K0}^T x^{0} ) x^{0} W_{V0} \\  \label{eq:two-stream} &\text{~~~~~~~~~}+ f( x^{1}W_{Q1} W_{K1}^T y^{0} ) y^{0}W_{V1} . 
\end{align}
In comparison, if we denote by $\phi^0$ the concatenated sequence of image and text tokens, our MAAF one-stream attention is given by
\begin{align}
    \phi^1 = \phi^0 + f(\phi^0 W_Q W_K^T \phi^0) \phi^0W_V .
\end{align}
In particular, the image tokens in the output of MAAF are 
\begin{align}\label{eq:one-stream}
     x^\text{one-stream} = x^0 &+ \frac{1}{A}\left[\exp(x^0 W_Q W_K^T x^0)x^0W_V \right] \\ \nonumber &+  \frac{1}{A}\left[\exp(x^0 W_Q W_K^T y^0)y^0 W_V)  \right]
\end{align}
where $$A=\sum_j \exp(x^0 W_Q W_K^T x^0_j) + \sum_j \exp(x^0 W_Q W_K^T y^0_j)$$ is the denominator of the softmax $f$. Rewriting the two-stream operation (\ref{eq:two-stream}) more explicitly for comparison with (\ref{eq:one-stream}),
\begin{align}
    x^\text{two-stream} = x^0 &+ \frac{1}{B}\exp( x^{0}W_{Q0} W_{K0}^T x^{0} ) x^{0}W_{V0} \\ \nonumber  &+ \frac{1}{C}\exp( x^{1}W_{Q1} W_{K1}^T y^{0} ) y^{0}W_{V1}
\end{align}
where $B$ and $C$ are softmax denominators, we can see that both operations include the same basic $f(xx)x$ and $f(xy)y$ interactions of the inputs $x,y$ (up to constants and learned matrices).

However, these equations also elucidate three potentially important differences between the two operations. First, in the two-stream operation the attention weights to the text tokens depend on $x^1$, which has context from the other image tokens. Second, the softmax is computed over the image-image and image-text attention weights separately in the two-stream case. Third, the two-stream case has an additional set of learnable parameters $W_{Q1}, W_{K1}, W_{V1}$. 

In our implementation of the two-stream model as two sublayers, there are also additional LayerNorm and Dropout operations which could have an effect. We also use multiple attention heads in each sublayer. The position-wise feedforward layers, however, are only applied after the operations in the equations above.

Which model would perform better is ultimately an empirical question, but our intuition favored the one-stream model. The one-stream model allows the same basic interactions with fewer parameters, and any benefit from using the post-self-attention $x^1$ for computing image-text attention weights could likely also be obtained from multiple blocks of the one-stream operation.

\section{Performance by type of modifier}
To better understand the source of our model's performance gains, we evaluated it on subsets of the Fashion IQ validation set that included keywords of several types, \textit{e.g.}, color (red, blue, green, ...) and texture (stripe, solid, plaid, ...). Table~\ref{table:fiq_by_modifier_type} shows the results of this experiment. These modifier classes are not mutually exclusive since a caption like ``is plaid blue and black" contains keywords for multiple modifier classes (in this case ``plaid" texture and ``blue and black" color). For all subsets, our model outperforms TIRG by roughly 16\% relative on (R10 + R50)/2.

\begin{table*}
\centering
\begin{tabular}{| l | c | c | c | c | c | c |}
\hline
 & all & color & texture & shape & parts & type \\
\hline
\hline
N & 6016 & 4902 & 1496& 3011 & 2906 & 313 \\
MAAF (Ours) & 36.86 & 39.18 & 41.74 & 36.23 & 38.44 & 37.89 \\
TIRG & 31.29 & 33.07 & 34.75 & 30.35 & 31.91 & 31.07 \\
Rel $\Delta$ & +15.1 & +15.6 & +16.7 & +16.2 & +17.0 & +18.0 \\
\hline
\end{tabular}
\caption{Fashion IQ validation set Recall at 1, by rough classification of the types of modifier in the relative captions. \label{table:fiq_by_modifier_type}}
\end{table*}

\section{Visualization of Attention Weights on Fashion IQ}

Figure 6 in the main text displays the same effects that we have identified and explored in the CSS dataset, but in the less controlled setting of FashionIQ. For words in this dataset that correspond to specific parts of a clothing item, such as ``short sleeves", or ``longer/shorter dress", we notice that several attention heads highlight the corresponding parts of the image. More interestingly, for additive changes such as ``longer dress", the model shows large attention weights over the corresponding image region; whereas for subtractive changes such as ``shorter dress" the model shows small attention weights. Large and small attention weights here are to be understood in comparison to the attention weights for a randomly chosen word from the modifying caption.

For each word and modifying sentence in Figure 6 in the main text, we display the attention map averaged over multiple images in the test set.  To minimize the noise due to the variance of different clothing items' absolute positions in the images, we clustered the grayscale images in pixel-space (using $k$-means with $k=5$) and only consider the images in the largest cluster. The rescaled 14x14 attention weights computed for the first 200 images in that cluster, are overlaid on the average of those images. These average images can be seen in the third row of Figure \ref{fig:average_image} below. 
The variation in clothing item position over the first 10 images in the largest cluster (row 2) is visibly smaller than that over the first 10 images of the whole dataset (row 1). This observation suggests that averaging attention maps for the same word over multiple examples in the same cluster could give a more consistent, spatially localized map.

\begin{figure}
\begin{center}
\includegraphics[width=0.99\linewidth]{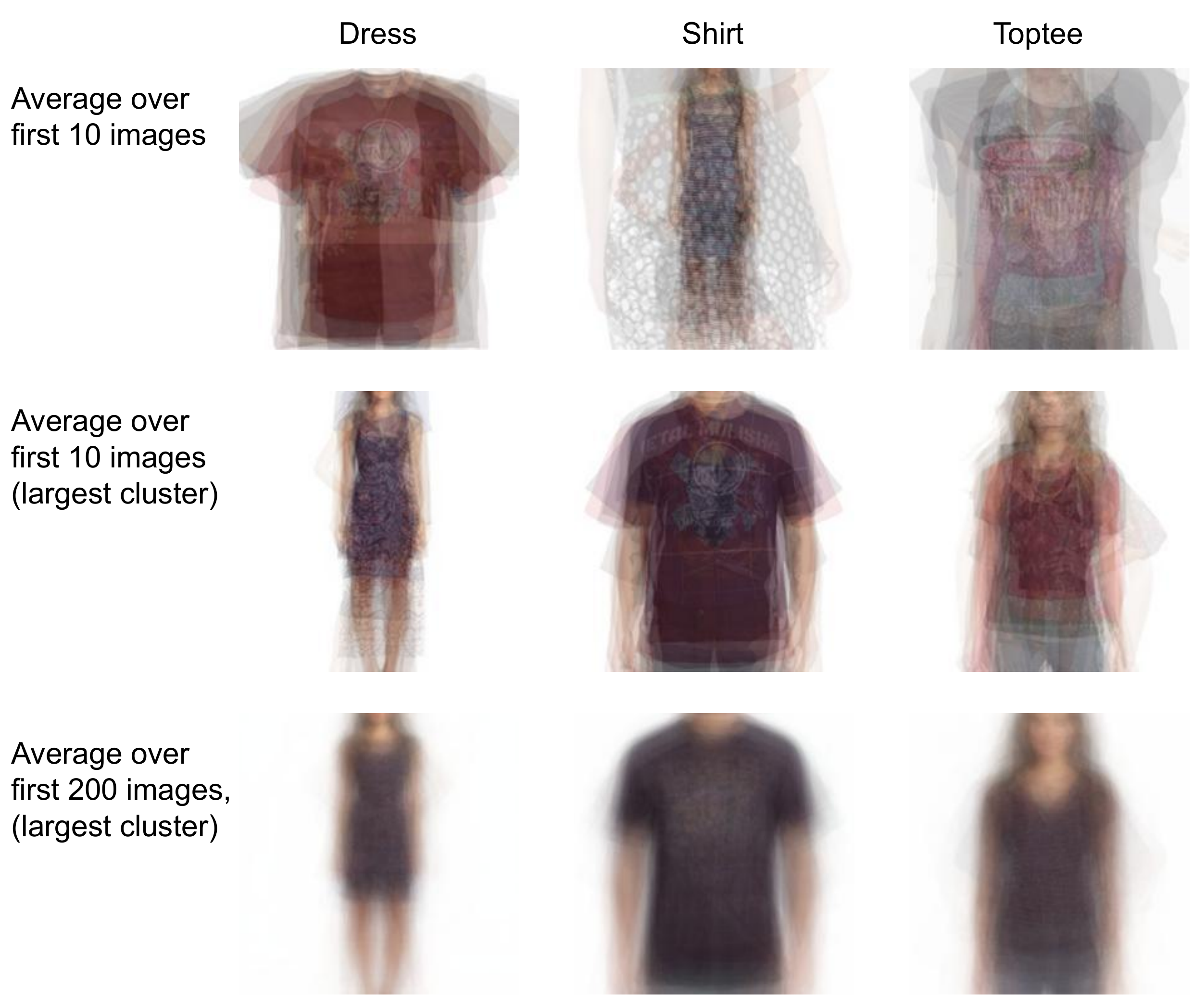}
\end{center}
   \caption{Average image for each of the Fashion IQ clothing categories, over the whole dataset or the largest cluster in pixel space.}
\label{fig:average_image}
\end{figure}

\section{Retrieval examples on Fashion IQ}
\begin{figure*}
\begin{center}
\includegraphics[width=0.99\textwidth]{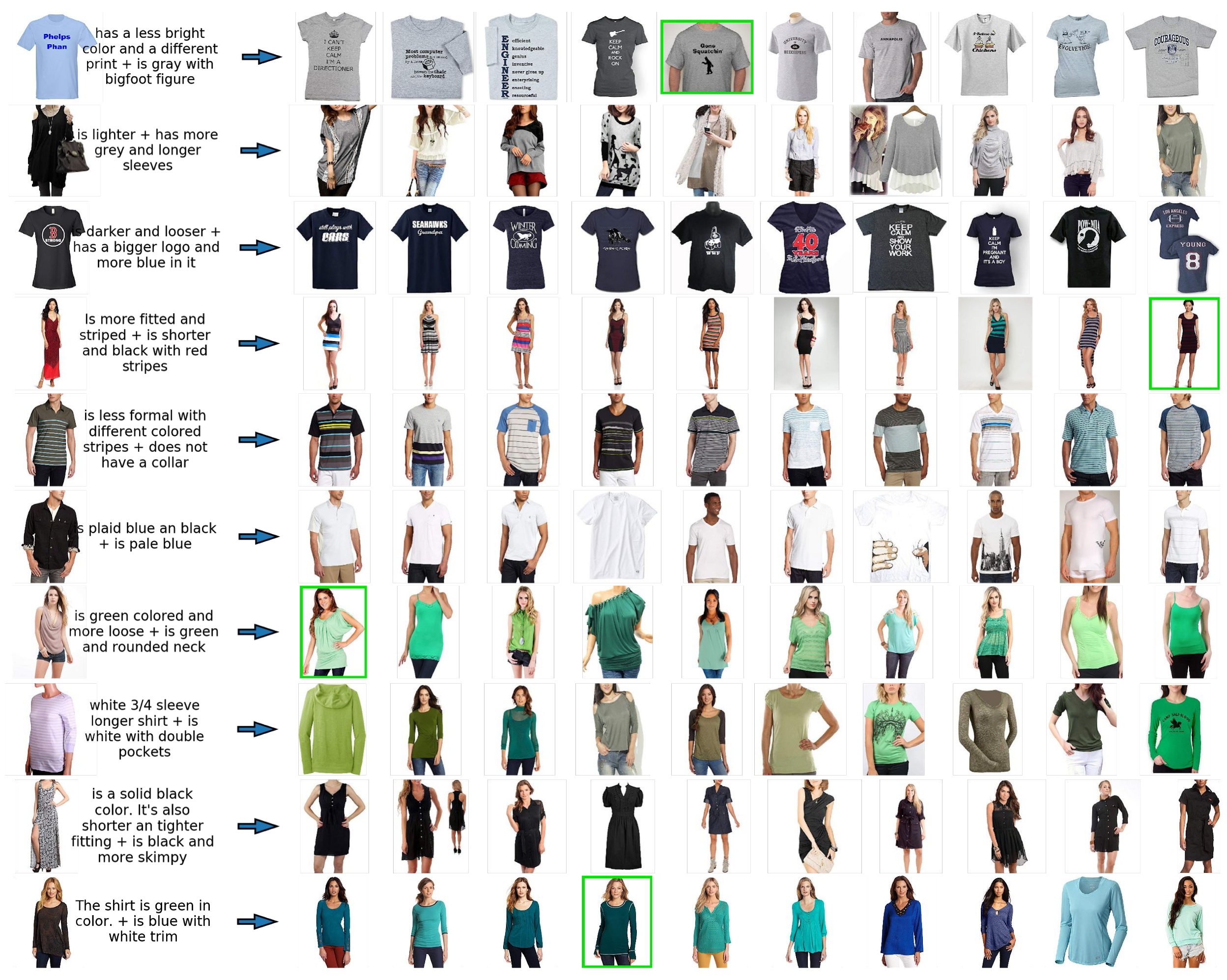}
\end{center}
   \caption{Example retrievals on the Fashion IQ validation set. A green rectangle indicates the labeled ``target" for the query.}
\label{fig:more_fiq_examples}
\end{figure*}

Figure 1 in the main text displays the top 3 retrievals using MAAF for several queries in the Fashion IQ validation set. Figure \ref{fig:more_fiq_examples} displays the top 10 retrievals for 10 more randomly selected queries from the Fashion IQ validation set. Note that the dataset does not contain exhaustive labels, so some appropriate retrievals are not labeled as correct.

The two captions provided for each query are displayed joined by ``+" and without modification. The captions are presented in this manner to the model but with punctuation removed and all letters lowercase. We made various efforts to normalize the text by spell checking, lemmatization, and stopword omission, but in early experiments these steps did not improve the performance of our models and so were not used for the experiments in the paper.

\section{MIT States}

The MIT States dataset \cite{isola2015discovering} consists of about 60k images, each with an object/noun label and a state/adjective label. To compare with \cite{vo2019composing}, we evaluate image retrieval using queries of an image together with the state/adjective for a target image with the same object/noun. The nouns in the test set are not seen during training, so the system needs to apply adjectives it knows to unseen objects. Table \ref{table:mit_states} shows our results compared to other existing models.

\begin{table*}
\begin{center}
\begin{tabular}
{|l|c|c|c|}
\hline
Method & R1 & R5 & R10 \\
\hline
\hline
Show \& Tell~\cite{vinyals2015show} & 11.9 $\pm$ 0.1 & 31.0 $\pm$ 0.5 & 42.0 $\pm$ 0.8 \\
Att. as Oper.~\cite{nagarajan2018attributes} & 8.8 $\pm$0.1 & 27.3 $\pm$ 0.3 & 39.1 $\pm$0.3\\
FiLM~\cite{perez2018film} & 10.1 $\pm$ 0.3 & 27.7 $\pm$ 0.7 & 38.3 $\pm$ 0.7 \\
TIRG~\cite{vo2019composing} & 12.2 $\pm$ 0.4 & 31.9 $\pm$ 0.3 & 43.1 $\pm$ 0.3 \\
Relationship~\cite{santoro2017simple} & 12.3 $\pm$ 0.5 & 31.9 $\pm$ 0.7 & 42.9 $\pm$ 0.9 \\
Locally Bounded~\cite{Hosseinzadeh_2020_CVPR} & \textbf{14.7} $\pm$ 0.6 & \textbf{35.3} $\pm$ 0.7 & \textbf{46.6} $\pm$ 0.5 \\
MAAF (Ours) & 12.7 $\pm$ 0.8 & 32.6 $\pm$ 0.5 & 44.8 $\pm$ 0.9 \\ %
\hline
\end{tabular}
\end{center}
\caption{Our results on MIT States, with comparisons to existing methods as reported in \cite{vo2019composing} and \cite{Hosseinzadeh_2020_CVPR}. \label{table:mit_states}}
\end{table*}

\end{document}